\documentclass[letterpaper]{article} % DO NOT CHANGE THIS
\usepackage{aaai25}  % DO NOT CHANGE THIS
\usepackage{times}  % DO NOT CHANGE THIS
\usepackage{helvet}  % DO NOT CHANGE THIS
\usepackage{courier}  % DO NOT CHANGE THIS
\usepackage[hyphens]{url}  % DO NOT CHANGE THIS
\usepackage{graphicx} % DO NOT CHANGE THIS
\usepackage{natbib}  % DO NOT CHANGE THIS AND DO NOT ADD ANY OPTIONS TO IT
\usepackage{caption} % DO NOT CHANGE THIS AND DO NOT ADD ANY OPTIONS TO IT
\usepackage{algorithm}
\usepackage{algorithmicx}
\usepackage{algpseudocode}
\usepackage{newfloat}
\usepackage{listings}
\DeclareCaptionStyle{ruled}{labelfont=normalfont,labelsep=colon,strut=off} % DO NOT CHANGE THIS
\floatstyle{ruled}
\newfloat{listing}{tb}{lst}{}
\floatname{listing}{Listing}
\usepackage{diagbox}
\usepackage[utf8]{inputenc} % allow utf-8 input
\usepackage{url}            % simple URL typesetting
\usepackage{booktabs}       % professional-quality tables
\usepackage{amsfonts}       % blackboard math symbols
\usepackage{nicefrac}       % compact symbols for 1/2, etc.
\usepackage{microtype}      % microtypography
\usepackage{xcolor}         % colors
\usepackage{subfigure}

\usepackage{graphicx}
\usepackage{amsmath}
\usepackage[utf8]{inputenc} % allow utf-8 input
\usepackage{url}            % simple URL typesetting
\usepackage{booktabs}       % professional-quality tables
\usepackage{amsfonts}       % blackboard math symbols
\usepackage{nicefrac}       % compact symbols for 1/2, etc.
\usepackage{microtype}      % microtypography
\usepackage{xcolor}         % colors
\usepackage{array}
\usepackage{bm}

\usepackage{microtype}
\usepackage{graphicx}
\usepackage{subfigure}
\usepackage{booktabs} % for professional tables

\usepackage{amssymb}
\usepackage{mathtools}
\usepackage{amsthm}

\usepackage{multirow}

\usepackage[capitalize,noabbrev]{cleveref}
\usepackage{diagbox}

\title{Sequential Preference Optimization: Multi-Dimensional Preference Sequential Alignment With Implicit Reward Modeling}
\author{
    Xingzhou Lou$^{1,2}$, Junge Zhang$^{1,2}$\thanks{Correspondence}, Jian Xie$^3$, Lifeng Liu$^3$, Dong Yan$^3$, Kaiqi Huang$^{1,2}$
}
\affiliations{
    \textsuperscript{\rm 1}Institute of Automation, Chinese Academy of Sciences\\
    \textsuperscript{\rm 2}School of Artificial Intelligence, University of Chinese Academy of Sciences\\
    \textsuperscript{\rm 3}Baichuan Inc.\\
    \texttt{louxingzhou2020@ia.ac.cn, jgzhang@nlpr.ia.ac.cn, xiejian1990@gmail.com,} \\
    \texttt{sproblvem@gmail.com, liulifeng@baichuan-inc.com,kqhuang@nlpr.ia.ac.cn}
}

\usepackage{bibentry}
\begin{document}

\maketitle

\begin{abstract}
Human preference alignment is critical in building powerful and reliable large language models (LLMs). However, current methods either ignore the multi-dimensionality of human preferences (e.g. helpfulness and harmlessness) or struggle with the complexity of managing multiple reward models. To address these issues, we propose Sequential Preference Optimization (SPO), a method that sequentially fine-tunes LLMs to align with multiple dimensions of human preferences. SPO avoids explicit reward modeling, directly optimizing the models to align with nuanced human preferences. We theoretically derive closed-form optimal SPO policy and loss function. Gradient analysis is conducted to show how SPO manages to fine-tune the LLMs while maintaining alignment on previously optimized dimensions. Empirical results on LLMs of different size and multiple evaluation datasets demonstrate that SPO successfully aligns LLMs across multiple dimensions of human preferences and significantly outperforms the baselines.
\end{abstract}

\section{Introduction}
Pretrained large language models (LLM) like GPT-4 \cite{gpt4} and Llama \cite{touvron2023llama1,touvron2023llama,llama3} are trained on very large corpus of text and demonstrate surprising capabilities in multiple domains, such as natural language processing \cite{jiao2023chatgpt,singhal2023large}, programming \cite{nijkamp2022codegen,qian2023communicative} and decision making \cite{wang2023voyager,zhang2023building}. These models are fine-tuned with humans' feedback to align with certain human preferences, e.g. harmlessness and helpfulness. Human preference alignment improves LLM's ability to generate responses preferred by humans and is essential in building AI assistants \cite{gpt4,touvron2023llama,claude2,jiang2024mixtral,llama3}. Specifically, Reinforcement Learning with Human Feedback (RLHF) \cite{ouyang2022training} learns a reward model to discriminate preferred and less preferred responses, and then optimizes LLMs with the reward model and RL algorithms. Direct Preference Optimization (DPO) \cite{rafailov2023direct} omits fitting an explicit reward model and directly optimizes LLMs to adhere to human preferences, and thus is known as implicit reward modeling.

Prevalent preference alignment methods focus on fine-tuning LLMs based on ranked response pairs, which only indicate which response is generally better \cite{zheng2023judging,chiang2024chatbot}. However, instead of solely \emph{good} or \emph{bad}, texts usually have multi-dimensional properties. For instance, a concise text summary generated by LLMs may not be as informative as a relatively longer, but highly specific response. In this case, the concise response is preferred brevity-wise, while the specific response is preferred informativity-wise. In other words, preferences on different dimensions may contradict each other.

The most straightforward approach to deal with multi-dimensional preferences is to mix them into one single dimension to indicate which response is generally better. In this case, the alignment results could be significantly influenced by the annotators' subjective perception and ranking inconsistency across dimensions. Therefore, it is necessary to align LLMs on each dimension and strive a balance that accommodates preferences across all dimensions. Current methods \cite{jang2023personalized,dai2023safe} decouple preferences along dimensions and align LLMs on each of the dimension by RLHF. However, they demand a reward model for each dimension. Fine-tuning LLMs with one reward model is already notoriously challenging. Multiple reward models further exaggerate this issue.

To address these issues, we propose Sequential Preference Optimization (SPO) to align LLMs with multi-dimensional preferences in a sequential manner. Specifically, SPO incorporates multi-round fine-tuning, optimizing one specific preference dimension for each round. SPO adopts additional constraints to guarantee alignment on previous dimensions in the learning objective. Consequently, LLMs acquire the skill to align with one specific aspect of human preference in each round, while staying aligned with preferences in previous rounds. Also, SPO omits explicit reward modeling and directly optimizes preferences, thereby avoiding the issues of multiple reward models in RLHF-based methods.

Theoretically, closed-form optimal policy and loss function for SPO are derived. The loss function is a simple classification loss and can be optimized efficiently. Furthermore, we perform gradient analysis to illustrate how SPO effectively preserves alignment results of previous rounds of fine-tuning.

Empirically, we conduct experiments on the PKU-SafeRLHF-30k dataset \cite{ji2023beavertails}, where response pairs are separately ranked on the dimensions of helpfulness and harmlessness. Our experiments use Llama 2 7B and 13B \cite{touvron2023llama} as base models. Fine-tuned models are evaluated on multiple datasets \cite{alpaca_eval,bai2022training,ji2023beavertails}. We alsp include experiments with more preference dimensions. Results of these experiments suggest that SPO successfully aligns LLMs across multiple dimensions and outperform both baseline methods and open models. Main contributions of this paper are:

(1) We propose Sequential Preference Optimization (SPO), which is able to sequentially align LLMs on multi-dimensional preferences.

(2) We theoretically derive the learning objective of SPO, ensuring multi-dimensional preference alignment. Our gradient analysis elucidates the mechanism by which SPO accomplishes this objective.

(3) Empirical results with multiple training and evaluation datasets demonstrate that SPO successfully aligns LLMs with multi-dimensional human preferences.

\section{Preliminaries}
Supervised fine-tuned (SFT) model $\pi_0$, based on pretrained models and high-quality demonstrations, is the initial model of SPO, RLHF and other preference optimization methods.

For a response pair $(y_1,y_2)$ of prompt $x$, $y_1\succ y_2$ represents that $y_1$ is the preferred response by humans. The preferences are decided by some unknown latent reward function $r^*(x,y)$. The prevalent way to model human preference distribution is the Bradley-Terry (BT) model \cite{bradley1952rank}, given by
\begin{equation}
    p(y_1\succ y_2|x)=\frac{\exp\left(r^*(x,y_1)\right)}{\exp\left(r^*(x,y_1)\right)+\exp\left(r^*(x,y_2)\right)}.
\end{equation}

For RLHF \cite{ziegler2019fine,stiennon2020learning,ouyang2022training}, LLMs are optimized with a learned reward model $r_\psi$ and Proximal Policy Optimization (PPO) \cite{schulman2017proximal}. The learning objective is to maximize preference rewards, constrained by a Kullback–Leibler (KL) divergence constraint.

DPO eliminates the need for explicitly fitting a reward model and uses the model with its reference for implicit reward modeling. The loss function in DPO is
\begin{equation}
    \label{dpo_obj}
    \mathcal{L}_{\pi_\theta}=-\mathbb{E}_\mathcal{D}\left[\log\sigma\left(\beta\log\frac{\pi_\theta(y_w|x)}{\pi_{\text{ref}}(y_w|x)}-\beta\log\frac{\pi_\theta(y_l|x)}{\pi_{\text{ref}}(y_l|x)}\right)\right],
\end{equation}
where $\pi_{\text{ref}}$ is the reference model which is usually the SFT model, $\beta$ is a hyperparameter, $(x,y_w,y_l)$ is sampled from the dataset $\mathcal{D}$, $y_w$ and $y_l$ are the preferred and dispreferred response, i.e. $y_w\succ y_l$. Minimizing Eq. \ref{dpo_obj} will make the model prefer chosen response $y_w$ over rejected response $y_l$ and align with human preferences.

\section{Related Work}
\textbf{Large Language Models} Pretrained models \cite{brown2020language,touvron2023llama1,li2023textbooks,llama3} acquire extensive world knowledge through self-supervised pretraining on an extraordinarily large corpus of texts. While pretrained models are able to predict the next words in sentences, they are not suitable for direct application in downstream tasks. However, with Instruction Fine-Tuning \cite{sanh2021multitask,chung2022scaling,ouyang2022training}, these models are trained on task-specific data, allowing them to follow prompts and excel at specific tasks. Thus, fine-tuned models exhibit strong capabilities across various domains.

\textbf{Preference Alignment} To prevent LLMs from generating unsatisfactory, misleading or even harmful responses \cite{bai2022training,kocon2023chatgpt}, LLMs must align with human preferences. RLHF \cite{ouyang2022training} trains a reward model with ranked response pairs, where higher rewards indicates better alignment with human preference. RLHF uses PPO \cite{schulman2017proximal} to fine-tune the LLMs to generate responses with high rewards from the reward model. However, fine-tuning LLMs with explicit reward modeling is notoriously complex and difficult \cite{bai2022training}. DPO \cite{rafailov2023direct} proposes implicit reward modeling, which can be optimized with a simple classification loss and significantly simplifies the fine-tuning pipeline.

Safe RLHF \cite{dai2023safe} and RL from Personalized Human Feedback (RL$\mathcal{P}$HF) \cite{jang2023personalized} also study alignment on multi-dimensional human preferences. However, their use of multiple reward models for alignment significantly complicates and destabilizes the fine-tuning process. Multi-Objective REward learning (MORE) \cite{zeng2023diverse} proposes to learn a multi-objective reward model by aggregation of shared reward and multiple dimension-specific reward heads. But compared to SPO, although MORE learns dimension-specific rewards, it does not guarantee alignment on each dimension. Rewarded soups \cite{rame2024rewarded} merges LLMs aligned with different datasets and objectives to combine their strengths.

\section{Methodology}
\begin{figure*}[t]
    %\hsize=\textwidth
    \centering
    \includegraphics[width=.85\textwidth]{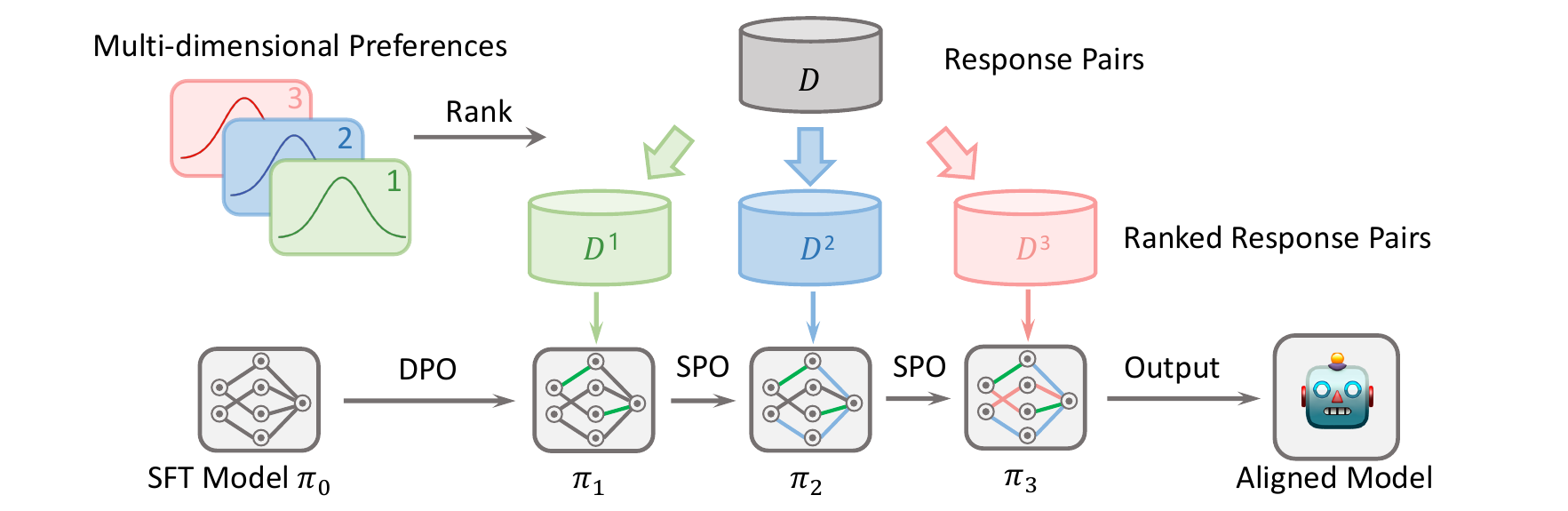}
    \caption{The SFT model is sequentially fine-tuned on multi-dimensional preferences with SPO, which aligns LLMs on the current dimension and preserves alignment on previous dimensions. First-round fine-tuning is achieved by DPO as there is no constraint on previous alignments.}
    \label{pipe}
    \vspace{-1em}
\end{figure*}

In this section, we will derive how to align LLMs with multi-dimensional human preferences in a sequential manner and propose Sequential Preference Optimization (SPO). We will first derive how to align LLMs on two-dimensional human preferences. Then, gradient analysis is conducted to show how SPO manages to achieve alignment across dimensions. Finally, we extend SPO to preference alignment with arbitrary number of dimensions. Pipeline of SPO is given in Fig. \ref{pipe}. 

To maintain previous alignment during sequential fine-tuning, rewards on previous dimensions must remain above a certain threshold. In SPO's pipeline, for the $n$-th round fine-tuning, $\pi_0$ and $\pi_{\{1,..,n-1\}}$ are the SFT model and previous sequentially fine-tuned models. The initial model $\pi_{n-1}$ maximizes $R_{n-1}$ and satisfy $\forall i \in \{1,..,n-2\},\ \mathbb{E}_{x\sim\mathcal{D},y\sim\pi_{n-1}}\left[R_i(x,y)\right]\geq H_i$, where $n\in\mathbb{N}\geq3$ and $H_i$ is the threshold for preference reward on the $i$-th dimension. In other words, $\pi_{n-1}$ is aligned with all previous dimensions.

To align with the $n$-th dimension and preserve previous alignment, the $n$-th round fine-tuning is formulated as
\begin{equation}
\begin{aligned}
         \max_{\pi_n}\mathbb{E}_{x\sim\mathcal{D},y\sim\pi_n}[R_n&(x,y)] \\
     s.t.\ \ \mathbb{D}_{KL}(\pi_n\|\pi_{n-1})\leq& H_{0}\\
           \forall\ i\in\{1,..,n-1\},\ -\mathbb{E}_{x\sim\mathcal{D},y\sim\pi_n}&\left[R_i(x,y)\right]\leq -H_i.
\end{aligned}
\label{optimization_problem}
\end{equation}
where $\mathcal{D}$ is the training dataset, $x$ is prompts from the dataset, $y$ is the response generated by $\pi_n$. The $n$-th round of fine-tuning in SPO ensures: (1) maximized reward for the $n$-th dimension, (2) limited deviation from $\pi_{n-1}$ and (3) prevention of significant degradation of previous alignments.

\subsection{Two-Dimensional Sequential Alignment}\label{two_round}
We first consider aligning LLMs on two-dimensional human preferences, i.e. $n=2$ in optimization problem Eq. \ref{optimization_problem}.

% Given an SFT model, the most straightforward way is to fine-tune the model on multi-dimensional preferences iteratively with off-the-shelf preference optimization algorithms \cite{ouyang2022training,rafailov2023direct}. Due to the \emph{alignment tax} \cite{ouyang2022training}, there is a performance cost for fine-tuning LLMs to align with human preferences. The alignment tax will accumulate in the iterative fine-tuning of each dimension, resulting in misalignment on previous dimensions or even catastrophic model collapse. To address these issues, SPO introduces additional constraints to the optimization problem to prevent model degradation.

\textbf{Deriving SPO Objective} Since there is no constraint on previous alignments in the first round fine-tuning, we can directly apply DPO on SFT model $\pi_0$ to obtain $\pi_1$, which maximizes preference reward $R_1$ on the first dimension.

The second round of fine-tuning in SPO solves Eq. \ref{optimization_problem} with $n=2$ and thus maximizes preference reward $R_2$ while preserving alignment on the first dimension. Like prior works \cite{peng2019advantage,rafailov2023direct}, we can derive the closed-form optimal policy $\pi_2^*$ for the constrained maximization problem Eq. \ref{optimization_problem}
\begin{equation}
    \label{eq:pi2*}
    \pi_2^*(y|x)=\frac{1}{Z_2(x)}\pi_1(y|x)\exp\left(\frac{\alpha_1}{\beta}R_1(x,y)+\frac{1}{\beta}R_2(x,y)\right),
\end{equation}
where $\beta$ controls deviation of $\pi_2$ from the reference model $\pi_1$, $\alpha_1$ controls the importance of maximizing reward $R_1$, and $Z_2(x)=\sum\limits_y\pi_1(y|x)\exp\left(\frac{\alpha_1}{\beta}R_1(x,y)+\frac{1}{\beta}R_2(x,y)\right)$ is the partition function. Detailed derivation of $\pi_2^*$ is given in the appendix A.1.

By taking logarithm on both sides and some algebra, Eq. \ref{eq:pi2*} can be transformed into
\begin{equation}
    R_2(x,y)=-\alpha_1 R_1(x,y)+\beta\log\frac{\pi_2(y|x)}{\pi_1(y|x)}+\beta\log Z_2(x),
\end{equation}
 where $R_1,R_2$ are based on BT model \cite{bradley1952rank}. $R_1=\beta\log\frac{\pi_1(y|x)}{\pi_0(y|x)}+\beta\log Z_1(x)$ can be represented by the SFT model $\pi_0$ and $\pi_1$ from the first round fine-tuning \cite{rafailov2023direct}. 

In BT model, preference is decided by the difference between responses' rewards. Specifically, $P_R(y_1\succ y_2)=\sigma\left(R(x,y_1)-R(x,y_2)\right)$, where $\sigma(x)=\frac{1}{1+e^{-x}}$ is the $sigmoid$ function. Thus, we can substitute $R_2$ into the BT model and derive the loss function for preference optimization on the second dimension, which is the log probability of preference in the BT model
\begin{equation}
    \begin{aligned}
    \label{eq:pre_trans_loss2}
    &\mathcal{L}_2^{SPO}(\pi^\theta_2)=-\mathbb{E}_{(x,y_w,y_l)\sim\mathcal{D}}\left[\log P_{R_2}(y_1\succ y_2)\right] \\
        &=-\mathbb{E}_\mathcal{D}\Bigg[\log\sigma\Bigg(\xi_2\phi_2(x,y_w,y_l)-\xi_1\phi_1(x,y_w,y_l)\Bigg)\Bigg],
    \end{aligned}
\end{equation}
where $\forall i\in\{1,2\},\ \phi_i(x,y_w,y_l)=\log\frac{\pi_i(y_w|x)}{\pi_{i-1}(y_w|x)}-\log\frac{\pi_i(y_l|x)}{\pi_{i-1}(y_l|x)}$, $x$ is the prompt, $y_w$ and $y_l$ are the preferred and the less preferred responses on the second dimension, $\mathcal{D}$ is the training dataset, constant $\xi_1=\alpha_1\beta>0$, $\xi_2=\beta>0$ and $\sigma$ is the $sigmoid$ function. $\phi_1$ is decided by the SFT model $\pi_0$ and the previous fine-tuned model $\pi_1$, while $\phi_2$ is decided by the current model $\pi_2$ and $\pi_1$. Detailed derivation of Eq. \ref{eq:pre_trans_loss2} is given in the appendix A.1. Minimizing $\mathcal{L}_2^{SPO}$ will lead $\pi_2$ to maximize human preference on the second dimension while still preserving preference alignment on the first dimension.

\textbf{Gradient Analysis} Compared to naive two-round sequential fine-tuning (where the constraint on $R_1$ is removed from the second round fine-tuning's optimization problem \ref{optimization_problem}), SPO is able to prevent the fine-tuned model from significant degradation on the preference maximization of $R_1$. We now theoretically explicate this advantage of SPO by analyzing gradient of the loss function $\mathcal{L}_2^{SPO}$.

The gradient of loss function $\mathcal{L}_2^{SPO}$ w.r.t. policy parameter of $\pi_2^\theta$ is given by
\begin{equation} 
\label{}
    \begin{aligned}
        \nabla_\theta\mathcal{L}_2^{SPO}&=-\xi_2\mathbb{E}_{\mathcal{D}}\Bigg[\sigma\big(-\xi_2\phi_2(x,y_w,y_l)+{\xi_1\phi_1(x,y_w,y_l)}\big)\\
        &\big(\nabla_\theta\log\pi_2^\theta(y_w|x)-\nabla_\theta\log\pi_2^\theta(y_l|x)\big)\Bigg].
    \end{aligned}
\end{equation}
Detailed derivation is given in appendix A.2. The first term $-\xi_2\phi_2(x,y_w,y_l)$ inside the $sigmoid$ function is for preference maximization on the second dimension. It assigns higher weight to the gradient when less preferred response $y_l$ has a high likelihood to be generated by $\pi_2$. This term will also appear in the gradient if we directly run DPO instead of SPO for the second round fine-tuning.

However, $\phi_1(x,y_w,y_l)=R_1(x,y_w)-R_1(x,y_l)=\log\frac{\pi_1(y_w|x)}{\pi_{0}(y_w|x)}-\log\frac{\pi_1(y_l|x)}{\pi_{0}(y_l|x)}$ in the second term is the reward difference between $y_w$ and $y_l$ given by reward $R_1$ on the first dimension. $\phi_1(x,y_w,y_1)>0$ when the preferred response $y_2$ on the second dimension is also preferred on the first dimension, and $\phi_1(x,y_w,y_l)<0$ when preferred responses are different on the two dimensions.

The weight of gradient increases when $\phi_1$ is positive (preferred responses are consistent) and decreases when $\phi_1$ is negative (preferred responses are inconsistent). Therefore, $\phi_1$ serves as a regularizer to prevent degradation of preference maximization on the first dimension. Consequently, SPO will strive to optimize the LLM so that preference maximization on both dimensions are achieved.
\subsection{Multi-Dimensional Sequential Alignment}
Now, we extend SPO to multi-dimensional preference alignment with arbitrary rounds of fine-tuning, i.e. $ n\in\mathbb{N},\ n\geq3$ in optimization problem Eq. \ref{optimization_problem}.

By solving the optimization problem, we have for $\forall n\in\mathbb{N}\geq3$, reward on the $n$-th preference dimension in SPO is
\begin{equation}
    \begin{aligned}\label{eq:r_multi}
        R_n(x,y)=\sum\limits_{i=1}^n\kappa_i\log\frac{\pi_i(y|x)}{\pi_{i-1}(y|x)},
    \end{aligned}
\end{equation}
where $\kappa_n=\beta$, $\kappa_{n-1}=-\beta\alpha_{n-1}$ and $\forall i\in\{2,..,n-1\},\ \kappa_{n-i}=-\beta\alpha_{n-i}\prod\limits^{i}_{j=2}(1-\alpha_{n-1-i+j})$ and $\alpha_k$ controls the importance of the $k$-th dimension. Detailed derivation is given in the appendix B.1.

\emph{Proof Sketch.} Similar as Eq. \ref{eq:pi2*}, we can first obtain the closed-form optimal solution of $\pi_n^*$, which is represented by the current reference model $\pi_{n-1}$ and preference rewards from previous rounds. By some algebra, we can get the formulation of $R_n$. Since the formulations of $R_1$ and $R_2$ are already at hand, we can iteratively substitute previous preference rewards into the formulation of $R_n$ and prove by mathematical induction that Eq. \ref{eq:r_multi} holds for $\forall n\geq3$. $\hfill\square$\begin{table*}[t]
\centering
\begin{tabular}{cccccccc}
\toprule
\multicolumn{1}{c}{Size}  & \multicolumn{1}{c}{Dataset} & \multicolumn{1}{c}{Safe-RLHF} & \multicolumn{1}{c}{DPO-Mix} & \multicolumn{1}{c}{RLHF-RS} &  \multicolumn{1}{c}{S-DPO} & DPO-HP& DPO-HM\\
\midrule
\multirow{4}{*}{7B} & HH-helpful     & {77.7\%}     & {87.3\%}            & 80.8\%                    & {81.5\%}              & {47.9\%}    & {64.6\%}      \\
                           & AlpacaEval     & 64.9\%            & 77.9\%                          & {67.6\%}                        & 78.0\%              &   47.7\%     &  {68.6\%}  \\
                           & HH-harmless     & {58.1\%}           & {57.7\%}        & 58.0\%                 & {41.9\%}                   & {63.3\%}    & {48.1\%}  \\
                           & SafeRLHF & 55.8\%           & 55.4\%                          & {56.5\%}                        & 40.5\%                    & 65.8\%  &  {46.3\%}  \\
\midrule
\multirow{4}{*}{13B} 
& HH-helpful        & {82.8\%}           & {59.9\%}                    & {85.3\%}                        & 81.5\%{}              & {47.2\%}       & {64.0\%}  \\
           & AlpacaEval      &    74.4\%           & {59.0\%}                 &          73.4\%              & {81.5\%}                                   & {44.1\%}     &   66.3\%   \\ 
            & HH-harmless     &      53.1\%      & {58.6\%}                        & {75.0\%}                      &        57.5\%            & {76.1\%}   & {51.8\%}  \\
            & SafeRLHF       &        51.3\%         & {60.0\%}           & {78.9\%}                       &        54.3\%           & {77.6\%}    & {51.0\%}    \\
\bottomrule
\end{tabular}
\caption{SPO's win rate against the baselines. The win rate is the proportion of questions where SPO gives better responses. HH-helpful and AlpacaEval evaluate LLMs' helpfulness while HH-harmless and SafeRLHF (short for PKU-SafeRLHF-Test) are evaluation datasets for harmlessness.}
\label{main_res}
\vspace{-1em}
\end{table*}

Specifically, if all previous dimensions are equally important, i.e. $\forall\ k\in\{1,..,n-1\}, \alpha_k=\alpha$, the preference reward is given by $\hat{R}_n=\beta\log\frac{\pi_n(y|x)}{\pi_{n-1}(y|x)}-\beta\sum\limits_{i=1}^{n-1}\alpha(1-\alpha)^{i-1}\log\frac{\pi_{n-i}(y|x)}{\pi_{n-i-1}(y|x)}$.

After obtaining the preference reward, SPO optimizes the LLM by directly maximizing the log probability of preference in the BT model. Loss function of the $n$-th round fine-tuning is given by
\begin{equation}\label{eq:loss_multi}
    \mathcal{L}_n^{SPO}(\pi_n^\theta)=-\mathbb{E}_{x,y_w,y_l\sim\mathcal{D}}\bigg[\sigma\big(R_n(x,y_w)-R_n(x,y_l)\big)\bigg].
\end{equation}
Optimizing $\pi_n$ by minimizing $\mathcal{L}_n^{SPO}$ enables the LLM to align with multi-dimensional preferences. Also, due to the constraints in our problem formulation, SPO is able to minimize the impact of alignment tax accumulated in multiple round of fine-tuning, achieving alignment across dimensions.

It is worth noting that the sequential fine-tuning in SPO only depends on the inference of previous models on the dataset $\mathcal{D}$, which has been done in the previous round of fine-tuning. Thus, no additional inference is required in the sequential training of SPO. We can just cache the inference results of previous rounds and use them in subsequent training. This makes the sequential fine-tuning in SPO very efficient.

We also conduct gradient analysis to theoretically demonstrate how SPO achieve multi-dimensional alignment. Details are given in appendix B.2. And pseudo code of SPO is given in the appendix C.

\section{Experiment}
In this section, we first evaluate SPO on a real-world dataset with two preference dimensions and then demonstrate SPO's ability to achieve multi-dimensional alignment with more preference dimensions.
\subsection{Experiment Setting}
\textbf{Training Datasets} Our training dataset for two-dimensional tasks is PKU-SafeRLHF-30k \cite{ji2023beavertails}, consisting of 26.9k response pairs ranked separately on the dimensions of helpfulness and harmlessness. Besides the original dataset, we also adopt a dataset altered from PKU-SafeRLHF-30k, designed to increase the challenge of multi-dimensional alignment. This modified dataset creates a contradiction between the two dimensions by including completely harmless but unhelpful responses, i.e. refusals. More details are provided in the appendix D.1. Our real-world training dataset for multi-dimensional preference is Helpsteer2 \cite{wang2024helpsteer2}, which consists of 10.1k response pairs annotated on multiple dimensions, e.g. helpfulness, coherence and verbosity.

\textbf{Evaluation Datasets} Four test datasets are employed for evaluation. HH-helpful \cite{bai2022training} (2.3k questions) and AlpacaEval \cite{alpaca_eval} (805 questions) are adopted in the evaluation of alignment on helpfulness. And we use HH-harmless \cite{bai2022training} (2.3k questions) and 300 random questions from PKU-SafeRLHF-test \cite{ji2023beavertails} to evaluate the alignment on harmlessness. 

\textbf{Models} The base model in our experiments is Llama 2 \cite{touvron2023llama}. SFT models are obtained by training the base models on Alpaca dataset \cite{alpaca} with supervised fine-tuning for 2 epochs. Llama 2 with 7B and 13B parameters are adopted to test the scalability of SPO.

\textbf{Baselines} Our baselines include (1) \textbf{Safe-RLHF} \cite{dai2023safe}. Safe RLHF achieves multi-dimensional alignment by learning a reward model for each dimension and safe RL algorithms (PPO-Lagragian \cite{ray2019benchmarking}); (2) RLHF with reward shaping (\textbf{RLHF-RS}) \cite{dai2023safe}; (3) \textbf{DPO-Mix}, where we mix the two-dimensional preferences into one dimension and run DPO on the mixed dataset. The principle of mixing is to prioritize harmlessness over helpfulness. Ranking for pairs with two harmless responses is randomly decided; (4) Sequential DPO (\textbf{S-DPO}). S-DPO is an ablation of SPO that sequentially runs DPO on each dimension without considering previous alignments; (5) DPO-Helpful (\textbf{DPO-HP}). DPO-HP is only fine-tuned on the dimension of helpfulness; (6) DPO-Harmless (\textbf{DPO-HM}), which is only fine-tuned on the dimension of harmlessness.

\textbf{Training Details} We adopt LoRA adapters \cite{hu2021lora,zheng2024llamafactory} to achieve efficient fine-tuning. For DPO-based methods, we set $\beta=0.1$, LoRA rank to $8$, LoRA scaling factor to $32$, learning rate to $1\times10^{-5}$. For SPO, we set $\alpha=0.1$ and other hyperparameters are kept the same as DPO. For SPO and S-DPO, models are sequentially fine-tuned on both dimensions for 2 epochs. For Safe-RLHF and RLHF-RS, we keep all default settings in the official implementation. For experiments on the original training set, we first fine-tune on the dimension of harmlessness and then on helpfulness. But since the modified dataset can cause the model to overfit on harmlessness and lose linguistic diversity, we first fine-tune the models on helpfulness and then on harmlessness. 

\textbf{Evaluation Metric} After fine-tuning, we pair SPO's responses with baselines' responses and use LLM-as-a-judge \cite{zheng2023judging} to score their helpfulness, safety, intent understanding and quality of language. For helpfulness evaluation datasets, responses' final scores are calculated by averaging the scores except safety. And for harmlessness evaluation datasets, we exclude the helpfulness score. We report SPO's win rate (proportion of questions where SPO has better responses) as evaluation results. To overcome the positional bias of LLM evaluators \cite{wang2023large}, for each response pair, we score each response-pair twice with GPT-4, switching their positions each time, and average the results as the final score. Responses with higher final scores win. The prompt, win rate calculation and a human consistency study is given in the Appendix D.2 and D.3. \begin{figure*}[t]
    %\hsize=\textwidth
    \centering
    \includegraphics[width=.95\textwidth]{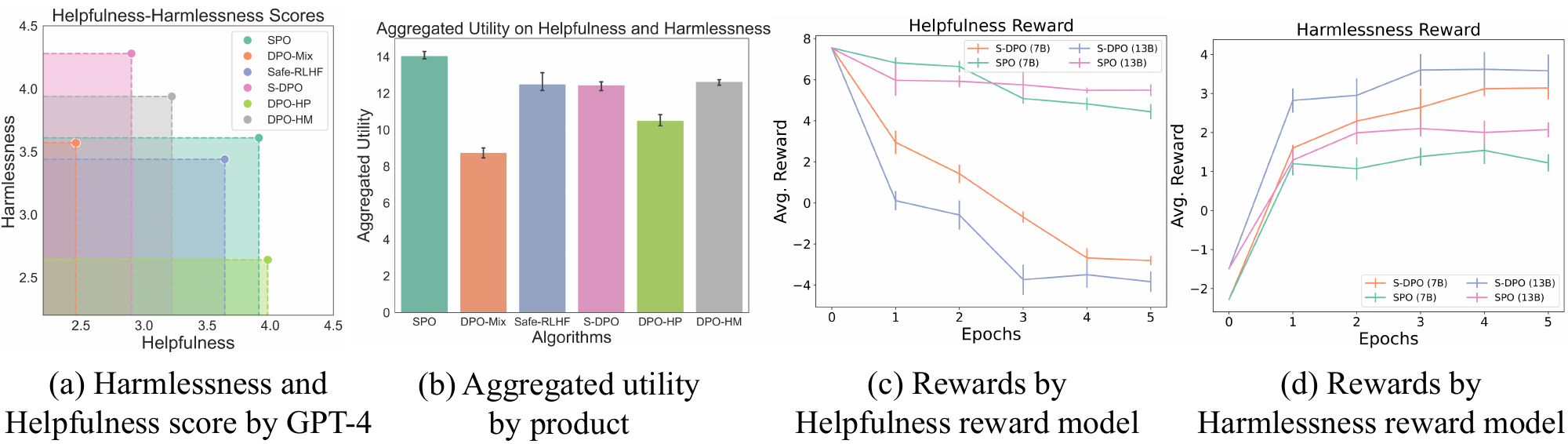}
    \caption{(a) Scores on harmlessness and helpfulness evaluation datasets by GPT-4 evaluator. (b) Aggregated utility of two dimensions, which is the product of harmlessness and helpfulness scores. (c), (d) give Helpfulness and Harmlessness rewards during the second-round fine-tuning process of SPO and S-DPO. SPO better preserves alignment on the first dimension of Helpfulness while learning to align with Harmlessness.}
    \label{square}
    \vspace{-1em}
\end{figure*} \begin{table*}[ht]
    \begin{minipage}[t]{0.49\linewidth}
    \centering
\begin{tabular}{cccccccc}
\toprule
\multicolumn{1}{c}{\diagbox[]{$R_i$}{$\alpha$}} &\multicolumn{1}{c}{$0$} & \multicolumn{1}{c}{$0.05$} & \multicolumn{1}{c}{$0.1$}  & \multicolumn{1}{c}{$0.3$} &  \multicolumn{1}{c}{$0.5$} &  \multicolumn{1}{c}{$0.7$} \\
\midrule
    Helpful &1.42& 6.53& 6.82& 6.93 & 7.05 & 7.13 \\
    Harmless &2.29& 1.30 & 1.18 & 0.22 & -0.74 & -1.42 \\
\bottomrule
\end{tabular}
\caption{The effect of hyperparameter $\alpha$ for preserving reward and alignment on previous dimensions.}
    \label{alpha_ablate}
    \end{minipage}\ \ \ \ \ \ \ 
    \begin{minipage}[t]{0.49\linewidth}
    \centering
	\begin{tabular}{cccccc}
        \toprule
        Dataset & S-DPO  & Alpaca & Vicuna & Mistral \\
        \midrule
	 HH-Helpful & 53.1\% &  54.6\%  & 55.7\%  & 49.8\% \\
      AlpacaEval & 56.3\% & 56.3\% & 51.5\% & 51.5\% \\
      HH-Harmless & 76.9\%  & 71.0\% & 55.5\% & 60.4\% \\
      SafeRLHF & 76.2\% & 60.8\% & 53.2\% & 48.8\% \\
        \bottomrule
	\end{tabular}
         \caption{SPO's win rate against open models.}
         \label{openmodels}
    \end{minipage}
    \vspace{-1.5em}
\end{table*}

\subsection{Results with the Modified Training Dataset}
We first give the experiment results after fine-tuning the models with our modified two-dimensional dataset, where alignment across dimensions is harder to achieve.

\textbf{Main Results} Table \ref{main_res} gives the evaluation win rate of SPO against the baselines. DPO-HP and DPO-HM perform slightly better than SPO on their fine-tuned dimension but significantly worse on the other dimension. The comparisons demonstrate the ability of SPO in reconciling the contradictions between these two dimensions and striking a compromise to achieve alignment on both dimensions. DPO-Mix has poor performance in all settings. This shows the importance of preference optimization for each dimension. S-DPO yields the same pipeline of SPO but has no constraints to preserve previous alignment results. For model with 7B parameters, as S-DPO has no additional constraints for preference optimization, it exhibits better alignment on the second dimension (harmlessness), but shows drastic degradation of alignment on the first dimension (helpfulness). For model with 13B parameters which has stronger expressive capability, S-DPO overfits to the second dimension of harmlessness, where it always gives extremely simple but harmless responses. Therefore, SPO exhibits a higher overall win rate even on the harmlessness evaluation datasets.\begin{table*}[ht]
    \begin{minipage}[t]{0.49\linewidth}
    \centering
\begin{tabular}{cccccc}
    \toprule
     & $\text{\emph{Token}}_1$& $\text{\emph{Token}}_2$&$\text{\emph{Token}}_3$&$\text{\emph{Token}}_4$ & Optimal             \\
    \midrule
    SPO & 98.0$\%$& 77.7$\%$& 64.7$\%$& 83.3$\%$ & 50.3$\%$ \\
    S-DPO & 10.7$\%$& 1.3$\%$& 0 & 99.7$\%$ & 0 \\
    \bottomrule
  \end{tabular}
    \caption{Percentage of presence of four special tokens in the responses and Pareto-optimal responses.}
      \label{toy_res}
    \end{minipage}\ \ \ \ \ \ \ 
    \begin{minipage}[t]{0.49\linewidth}
    \centering
\begin{tabular}{ccccc}
    \toprule
     & Helpful & Correct & Coherent & Verbose                \\
    \midrule
    S-DPO & 57.3$\%$ &	55.3$\%$	& 57.6$\%$	&49.4$\%$  \\
    Merged DPO & 66.2$\%$ &	67.7$\%$	& 73.5$\%$	&53.3$\%$  \\
    \bottomrule
  \end{tabular}
    \caption{SPO's win rate on each dimensions after fine-tuning on four dimensions on Helpsteer 2.}
      \label{helpsteer_res}
    \end{minipage}
    \vspace{-1em}
\end{table*} 

Compared to the RLHF-based counterparts SafeRLHF and RLHF-RS, SPO better aligns with preferences on helpfulness and harmlessness for both 7B and 13B models. The performance of RLHF-based methods shows that explicit reward modeling on multiple preference dimensions will destabilize the fine-tuning process, leading to sub-optimal performance. In contrast, SPO follows the implicit reward modeling as DPO \cite{rafailov2023direct} and achieves better preference alignment across all dimensions.

\textbf{Alignment Analysis} We visualize the alignment scores evaluated by GPT-4 in Fig. \ref{square}(a). The harmlessness scores are averaged scores on PKU-SafeRLHF-Test and the helpfulness scores are averaged scores on AlpacaEval with fine-tuned 7B models. From the results, we can see SPO has similar helpfulness scores as DPO-HP, which is only fine-tuned on the helpfulness dimension. This shows SPO's strong capability in preserving previous preference alignment results. Inspired by \cite{zheng2022ai} that conducts evaluation with two contradictory metrics, we use the product of helpfulness scores and harmless scores as the aggregated utility, which is also the area of the rectangles in Fig. \ref{square}(a). Fig. \ref{square}(b) gives the aggregated utility of each method. SPO strikes a balance between alignment on the two preference dimensions and thus has the highest aggregated utilities.

\textbf{Ablation Study} By setting hyperparameter $\alpha=0$, we remove the constraint on preserving previous alignment results and obtain S-DPO. Fig. \ref{square}(c), (d) gives the helpfulness and harmlessness rewards during the second-round fine-tuning. Results are obtained by querying Safe-RLHF's reward and cost models. Compared to SPO, S-DPO's helpfulness rewards significantly drops, which means severe degradation of alignment on helpfulness. Especially, 13B model's strong expressive capacity makes it rapidly overfit to the harmlessness dimension, resulting in poor alignment on helpfulness. In contrast, although SPO has relatively lower harmlessness scores, it effectively preserves previous alignment on helpfulness. As a result, SPO defeats S-DPO in terms of overall performance.

Then we set $\alpha$ in SPO during second-round fine-tuning of 7B models to different values to see its effect on multidimensional alignment. Larger $\alpha$ stands for greater importance of preserving previous alignment on helpfulness. Results are given in Table \ref{alpha_ablate}. When $\alpha=0$, previous alignment on helpfulness significantly deteriorates. As $\alpha$ increases, helpfulness rewards increases, implying better preservation of the first-round alignment. Conversely, harmlessness rewards decrease, as they contradict preferences on the first dimension. Thus, the result show the constraint in SPO is able to effectively preserve previous alignment. We also propose solving the dual problem of SPO to dynamically adjust $\alpha$ to maintain previous preference rewards near a specified threshold. For more details, please refer to Appendix F.

% \begin{figure}[h]
%     \centering
%     \includegraphics[width=.28\textwidth]{pics/overfitting.png}
%     \caption{Overfitting during second-round fine-tuning.}
%     \label{overfitting_fig}
% \end{figure}
 \begin{table}[t]
    \centering
\begin{tabular}{cccccc}
    \toprule
     \multicolumn{1}{c}{\diagbox[]{ }{$e$}} &\multicolumn{1}{c}{$1^{st}$} & \multicolumn{1}{c}{$2^{nd}$} & \multicolumn{1}{c}{$3^{rd}$}  & \multicolumn{1}{c}{$4^{th}$} &  \multicolumn{1}{c}{$5^{th}$} \\
    \midrule
    SPO & 0$\%$ &	0$\%$	& 0$\%$& 0$\%$& 0$\%$	 \\
    S-DPO & 62.5$\%$ &	30.5$\%$	& 41.5$\%$& 45.5$\%$& 46.0$\%$	 \\
    \bottomrule
  \end{tabular}
    \caption{Percentage of overfitting during second-round fine-tuning. SPO shows no overfitting as epoch $e$ increases.}
      \label{overfitting_table}
    \end{table}
\textbf{Overfitting Study} The constraints on on prior dimensions in SPO force LLMs to retain previous alignments. Thus, they are also able to keep LLMs from overfitting to the current preference dimension. We analyzed overfitting during fine-tuning by evaluating SPO and S-DPO on 200 safe questions from AlpacaEval, where refusal to answer these safe questions indicates overfitting to the harmlessness preference dimension. To obtain the results, we filtered out responses containing key words like “sorry”, “as an AI assistant” and manually identifying refusals. We use 7B models in this experiment and fine-tune them on helpfulness for 2 epochs followed by 5 epochs on harmlessness.

Because the training dataset is altered to induce refusals (completely harmless but unhelpful), S-DPO demonstrates severe overfitting as shown in Table \ref{overfitting_table}. But with the constraint to preserve alignment on helpfulness, SPO does not overfit to the harmlessness dimension even after 5 epochs.

\textbf{Model Merging} Recently, model merging techniques \cite{rame2024rewarded,rame2024warp,rame2024warm} successfully merge different reward models and LLMs in the weight space and combine the strengths of them. Here we study whether model merging technique is able to achieve alignment across multiple potentially conflicting dimensions.

We first merge the helpful RM and harmless RM in our experiment by linear interpolation with equal weights and then evaluate the RMs on held-out validation sets on both dimensions. Results in Table \ref{merge_rm} show that the merged RM perform poorly on both helpfulness and harmlessness due to the inherently conflicting goals of the models being merged.
\begin{table}[ht]
    \centering
	\begin{tabular}{ccc}
        \toprule
         & Helpful Val.  & Harmless Val.  \\
        \midrule
	 Helpful RM & 67.1\% &  37.1\% \\
      Harmless RM & 35.4\% & 71.0\%\\
      Merged RM & 51.2\%  & 48.2\%\\
        \bottomrule
	\end{tabular}
         \caption{Accuracy of the RMs on validation sets. A prediction is correct when RM gives higher reward to the preferred response than the dispreferred response.}
         \label{merge_rm}
         \vspace{-1em}
\end{table}

Then, we merge two LLMs aligned with helpfulness and harmlessness by DPO separately and evaluate the merged model against SPO. SPO's win rate against the merged DPO model is $\textbf{79.0\%}$ on AlpacaEval and $\textbf{41.2\%}$ on PKU-SafeRLHF-Test. We can tell that SPO significantly outperforms merged DPO on helpfulness but loses on harmlessness, showing the harmless LLM becomes dominant in the merged model. This is potentially because harmless responses exhibit simpler patterns than helpful responses, e.g. refusals. Thus, merging inherently conflicting LLMs is not an ideal choice to achieve multi-dimensional alignment, as some LLMs may easily become dominant over the others. 

\subsection{Results with the Original Training Dateset}
We now give the results when fine-tuning Llama 2 7B model with the original PKU-SafeRLHF-30k dataset and compare our model with some prevalent open models.

The open models we consider here are Alpaca \cite{alpaca}, Vicuna-7B-v1.5 \cite{zheng2023judging} and Mistral-7B-Instruct-v0.1 \cite{jiang2023mistral}, which is based on a stronger base model than Llama 2 used in SPO \cite{mistral7b}. As shown in Table \ref{openmodels}, SPO outperforms the ablation S-DPO, Alpaca and Vicuna. Although Mistral-Instruct's base model is significantly stronger, SPO is still able to achieve comparable results. This experiment further demonstrates SPO is able to align LLMs with multi-dimensional preferences and achieve strong performance.
\subsection{Experiments with More Preference Dimensions}\label{toy_exp_section}
To evaluate SPO's ability to achieve multi-dimensional alignment, we first conduct experiments on a demonstrative dataset and then give results on the real-world dataset Helpsteer2 \cite{wang2024helpsteer2}, both with four preference dimensions.

\textbf{Demonstrative Experiments} We randomly sample 10k samples from the training dataset and augment them with four special tokens, denoted as $\{[\text{\emph{Token}}_1],[\text{\emph{Token}}_2],[\text{\emph{Token}}_3],[\text{\emph{Token}}_4]\}$, to indicate preference. Specifically, the ranking on each dimension is determined by the presence of a unique token. On each dimension, a special token is added to the preferred response, and other tokens have $10\%$ probability to be added to both preferred and dispreferred responses. In this way, alignment on each dimension is indicated by presence of the corresponding special token. The Pareto-optimal model that aligns with four dimensions will always include all special tokens in the generations. SPO and S-DPO are sequentially fine-tuned on four dimensions for 1 epoch. Please refer to Appendix E.1 for details.

From the results, we can see that after fine-tuning on 4 dimensions, SPO achieves multi-dimensional alignment with 50.3$\%$ Pareto-optimal responses (all special tokens are present). However, S-DPO, the ablation of SPO without the constraints on previous alignments, only aligns with the last dimension and severely degrades on previous dimensions. Thus, the constraints in SPO are able to align LLMs with more preference dimensions, underscoring the effectiveness of SPO in achieving multi-dimensional alignments.

\textbf{Real-World Dataset} Helpsteer2 \cite{wang2024helpsteer2} consists of 10.1k response pairs annotated on multiple dimensions. We run SPO on the dimensions of helpfulness, correctness, coherence and verbosity sequentially and evaluate against S-DPO and a merged model from four different DPO models separately aligned on each dimension. We evaluate the models on Helpsteer2 validation set and report the win rate of SPO on different dimensions respectively. More details and prompt for evaluation are given in appendix E.2. Win rates of SPO on Helpsteer 2 are given in Table \ref{helpsteer_res}.

The results confirm SPO’s effectiveness in achieving multi-dimensional alignment. Specially, S-DPO only performs slightly better on the last dimension as it has no additional constraints. And merged DPO performs poorly in all dimensions because the models to be merged are trained on the same data with potentially conflicting preferences, affecting the overall performance after model merging.

\section{Conclusion and Limitation}\label{conclusion}
In this paper, we tackle the problem of aligning LLMs with multi-dimensional preferences and propose Sequential Preference Optimization (SPO). SPO avoids explicit reward modeling in RLHF and achieve multi-dimensional alignment by iteratively solving constrained optimization problems. The constrained optimization problem enables SPO to optimize preference on new dimensions while preserving the alignment in previous rounds. Theoretically, we derive the learning objective of arbitrary rounds of preference alignment in SPO and conduct gradient analysis to illustrate how SPO achieves alignment across dimensions. Empirically, extensive experiments and studies on different training datasets, evaluation datasets and preference dimensions confirm the efficacy of SPO in aligning LLMs across multiple dimensions.

The limitation of this work is although 7B and 13B models are considered, we do not include extremely large open models  \cite{adler2024nemotron,llama3} in our experiments due to computational limit. In the future, we plan to apply SPO on these large models and evaluate against state-of-the-art LLMs. Another promising direction is to introduce SPO to the iterative setting \cite{yuan2024self,guo2024direct} as they significantly outperform the original offline DPO. But since the iterative DPO is orthogonal to this paper, we leave it to future work.
% The limitation of this work is although we give theoretical results and a toy experiment with four preference dimensions, our empirical study on real-world dataset is limited to two preference dimensions. A future direction is to decompose current generally ranked datasets into more nuanced dimensions (such as \emph{3H} \cite{askell2021general}, helpfulness, harmlessness and honesty), and apply SPO to the more fine-grained datasets to improve preference alignments for LLMs.


\bibliography{aaai25}

\end{document}